\documentclass{article}

\usepackage{arxiv}
\usepackage{color}
\usepackage[T1]{fontenc}    
\usepackage{hyperref}       
\usepackage{url}            
\usepackage{booktabs}       
\usepackage{amsfonts}       
\usepackage{nicefrac}       
\usepackage{cite}
\usepackage{microtype}      
\usepackage{lipsum}
\usepackage{graphicx}
\usepackage{caption}
\usepackage{subfigure}
\usepackage{amsmath}
\usepackage{multirow}
\usepackage{array}
\usepackage{mathtools}
\usepackage{bbm}
\usepackage{algorithm}
\usepackage{algpseudocode}
\algtext*{EndWhile}
\algtext*{EndIf}
\algtext*{EndFor}
\usepackage[dvipsnames]{xcolor}
\definecolor{mypink}{RGB}{219, 48, 122}

\title{Clustering Based on Graph of Density Topology}
\newcommand{\printfnsymbol}[1]{
   \textsuperscript{\@*}
 }
\author {
      Zhangyang Gao \thanks{These authors contribute equally.}  \textsuperscript{ ,} \textsuperscript{\rm 1},
        Haitao Lin\footnotemark[1]  \textsuperscript{ ,} \textsuperscript{\rm 1},
        Stan. Z Li \thanks{Corresponding author.} \textsuperscript{ ,}    \textsuperscript{\rm 1} \\
     \textsuperscript{\rm 1} AI Lab, School of Engineering, Westlake University. \\
    $\{$gaozhangyang, linhaitao, stan.zq.li$\}$@westlake.edu.cn
}

\begin{document}
\maketitle

\begin{abstract}
Data clustering with uneven distribution in high level noise is challenging. Currently, HDBSCAN \cite{campello2013density,mcinnes2017accelerated} is considered as the SOTA algorithm for this problem. In this paper, we propose a novel clustering algorithm based on what we call {\em graph of density topology} (GDT). GDT jointly considers the local and global structures of data samples: firstly forming {\em local clusters} based on a density growing process with a strategy for properly noise handling as well as cluster boundary detection; and then estimating a GDT from relationship between local clusters in terms of a connectivity measure, giving {\em global topological graph}. The connectivity, measuring similarity between neighboring local clusters, is based on local clusters rather than individual points, ensuring its robustness to even very large noise. Evaluation results on both toy and real-world datasets show that GDT achieves the SOTA performance by far on almost all the popular datasets, and has a low time complexity of $\mathcal{O}(n\log{n})$. The code is available at {\color{red} https://github.com/gaozhangyang/DGC.git}. 
\end{abstract}

\section{Introduction}

Unsupervised clustering is a fundamental problem in machine learning, aimed to classify data points without labels into clusters. Numerous clustering methods including k-means   \cite{arthur2006k,sculley2010web}, spectral clustering  \cite{von2007tutorial,yan2009fast}, OPTICS  \cite{ankerst1999optics} and others   \cite{von2007tutorial,ester1996density,chazal2013persistence,frey2007clustering,comaniciu2002mean,schubert2017dbscan} have been proposed. However, clustering algorithms have been suffering from uneven distribution of data in high level noise, until HDBSCAN is proposed   \cite{campello2013density,mcinnes2017accelerated}. A key insight of HDBSCAN is based on the density clustering assumption: in an appropriate metric space, data points tend to form clusters in high-density areas whereas noise tends to appear in low-density areas. By dropping noise points and maximizing the stability of clustering, HDBSCAN has made a great advance in classifying samples into clusters. 
However, HDBSCAN (and other as well) has the following weaknesses: (1) It detects the global topological structure based on the connectivity defined on individual points with its sensitivity to bridge-like noise (seeing Fig. \ref{fig:comparision}) between two clusters. (2) During the clustering process, it may mistakenly classify true samples into noise. 

In this paper, we propose a novel algorithm, called graph of density topology (GDT), to solve the aforementioned.
GDT is able to detect local clusters and topological structure of the clusters from data and achieve robustness to high noise and diverse density distributions. Different from other clustering algorithms, GDT considers the local and global structure of the sample set jointly: firstly forming local clusters based on density growing process with a proper strategy for properly handling noise as well as detecting boundary points of local clusters, then establishing the global topological graph from relationship between local  clusters  in  terms  of  a  connectivity.  The  connectivity, measuring  similarity  between  neighboring  local  clusters, is based on local clusters rather than individual points, ensuring its robustness to even very large noise. Results of experiments on both toy and real-world datasets prove that GDT outperforms existing state-of-the-art unsupervised clustering algorithms by a large margin. Our contributions are summarized as follows:
\begin{itemize}
    \item We propose GDT, which is able to deal with data from uneven distribution at high noise levels.
    \item We evaluate GDT on different tasks, with performance superior to other unsupervised clustering algorithms.
    \item We accelerate GDT with its time complexity of $\mathcal{O}(n\log{n})$.
\end{itemize}
We provide the GDT code  at {\color{red}https://github.com/gaozhangyang/DGC.git}. 

The rest of this paper is organized as follows. In section 2, we introduce the motivation and preliminary knowledge of the paper. In section 3, we propose our method. In 3.1, local cluster detecting algorithm is proposed; Topo-graph construction and the method for pruning the weak edges are introduced in 3.2 and 3.3 respectively; Finally, we analyze the properties of our method in 3.4. Then in section 4, we show the results of the experiments on different datasets compared with other unsupervised clustering methods for classification and segmentation tasks. 

\section{Background}

In this section, we first introduce the notation and motivation of our work, with a simple example in 1-d case illustrating the relationships of 'density' in Fig. 1 (a), 'graph of density topology' in Fig. 1 (b) and (c). Then we give a simple guide on preliminary knowledge for density estimation and density growing process in our method.

\subsection{Notation and Motivation}
$\mathcal{X} = \{\boldsymbol{x}_i | i=1,2,\cdots,n\}$ is a set of samples in metric space $(\mathbb{R}^{d},d)$ and $f(\boldsymbol{x})$ is the density function on $\mathbb{R}^{d}$. A consensus of unsupervised clustering methods is that data points tend to form clusters in high-density areas, and noise points tend to appear in low-density areas. Therefore, based on the density function $f(\boldsymbol{x})$, points in $\mathcal{X}$ are separated into local clusters according to peaks of $f(\boldsymbol{x})$. Some clustering algorithms \cite{comaniciu2002mean} regard these local clusters as final results, but we assume that local clusters should not be independent. In this case, a topological graph $G = (V,E)$ is constructed, which is called \textit{graph of density topology} showing the connectivity strength among them, with vertex set $V = \{v_j| 1\leq j \leq m\}$, where $v_j$ represents a point set of $j$-th local cluster centered on $\boldsymbol{m}_j$, and the edge set $E = \{e_{i,j}| 1\leq i\leq m,1\leq j\leq m, i\neq j \}$, where $e_{i,j}$ represents the connectivity between $v_i$ and $v_j$. Points belonging to the same local cluster or strongly connected local clusters share the same label, otherwise they have different labels. For clustering tasks, we need to prune weak edges of $G$ to ensure the diversity of labels. Fig. 1 shows a simple example in 1-d case.
\begin{figure}[h]
    \centering
        \includegraphics[width=2.8in]{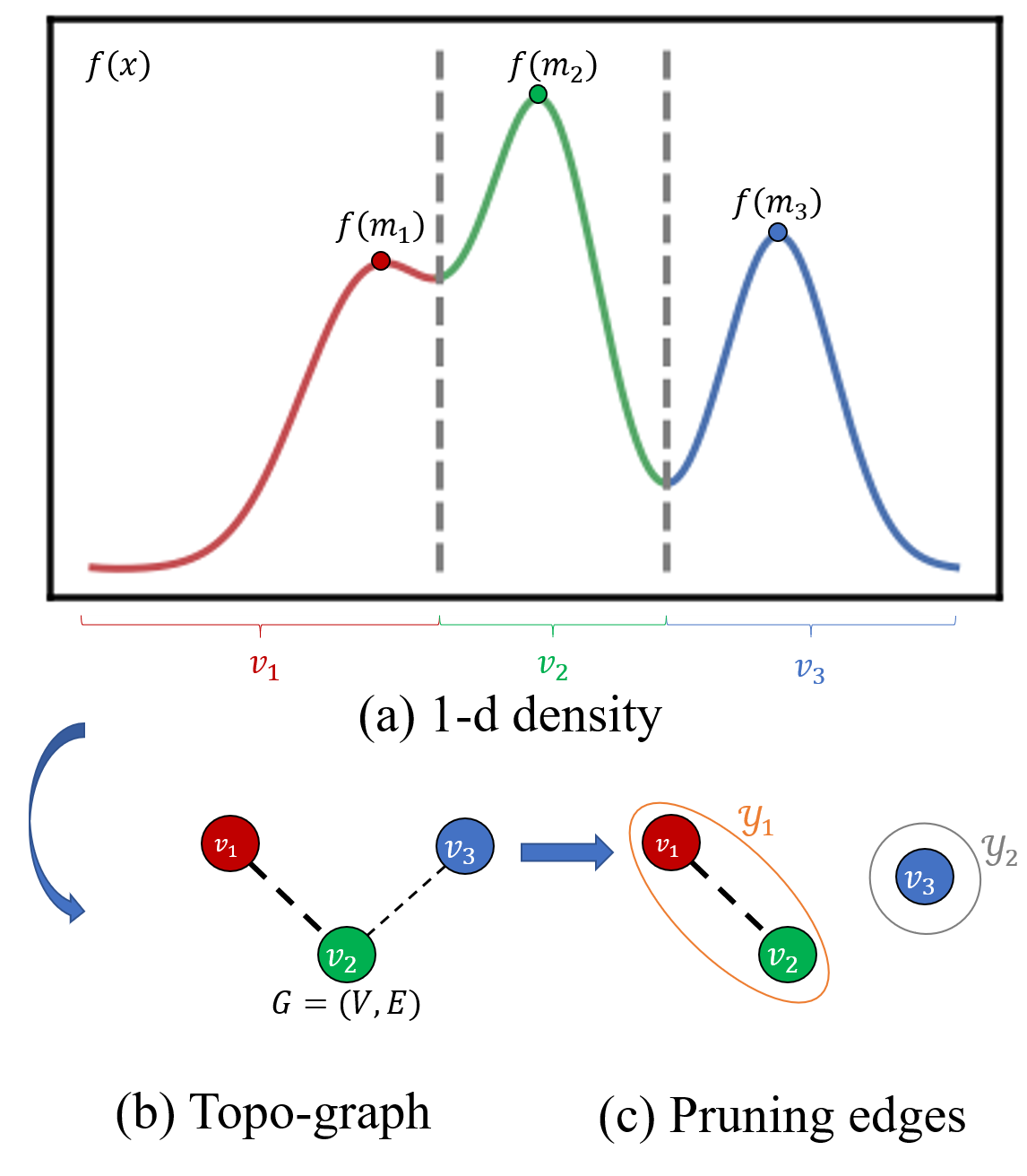}
	    \caption{The illustration of our motivation with 1d data: the estimated density is the full line in (a) with three peaks standing for three local clusters with centers: $m_1$, $m_2$ and $m_3$, colored with red, green and blue respectively. (b) shows the topological structure constructed by the density function, and after the weak edge ($e_{23}$) is pruned, we get the graph showed in (c), which indicates there are three local clusters with two shared labels $y_1,y_2$.}
	    \label{fig:motivation_1d}
\end{figure}
\subsection{Local Kernel Density Estimation}
\begin{figure}[ht]
	\centering
		\subfigure[Raw data]{ 
			\includegraphics[width=0.15\linewidth]{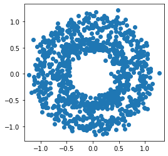}}
		\subfigure[LKDE]{ 
			\includegraphics[width=0.15\linewidth]{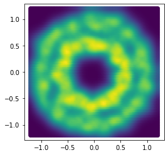}}
		\subfigure[KDE]{ 
			\includegraphics[width=0.15\linewidth]{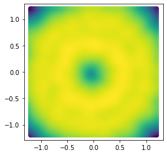}}\\
		\subfigure[LKDE(3-d)]{ 
			\includegraphics[width=0.20\linewidth]{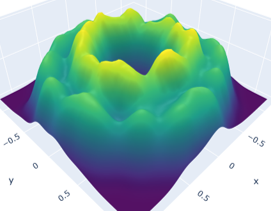}}
		\subfigure[KDE(3-d)]{ 
			\includegraphics[width=0.20\linewidth]{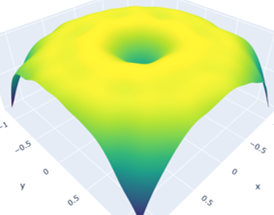}}
		\caption{The comparison of KDE and LKDE: (a) is raw data. (b) and (c) are the estimated densities via LKDE and KDE with Gaussian kernel respectively, shown in 2D plane. (d) and (e) are the 3d surface of (b) and (c), respectively. All the density functions have been normalized to $[0,1]$.}
		\label{fig:compre KDE and LKDE}
\end{figure}

Kernel density estimation(KDE) is a classical way to obtain the continuous density distribution of $\mathcal{X}$. However, due to the fat-tail characteristic of kernel functions and their sensitivity to bandwidth, the classical KDE often suffers from globally over-smoothing as shown in Fig \ref{fig:compre KDE and LKDE}. To avoid these shortcomings, local kernel density estimation(LKDE) is used in this paper, which can be formulated as
\begin{align}
\label{eq:LKDE}
    p(\boldsymbol{x}) = \sum_{\boldsymbol{x}' \in \mathcal{N}'_{\boldsymbol{x}}} \prod_{l=1}^d \kappa(x^{(l)} - x'^{(l)}) 
\end{align}
where $\boldsymbol{x'} = (x'^{(1)}, x'^{(2)},\cdots, x'^{(d)})^{\top}$ and $\mathcal{N}'_{\boldsymbol{x}}$ is $\boldsymbol{x}$'s neighbors for density estimation, $|\mathcal{N}'_{\boldsymbol{x}}|=k_d$ and $\kappa(\cdot) $ is kernel function. For stability, the density estimated by LKDE will be scaled by Max-min normalization 
\begin{align}
\label{eq:normalize}
    f(\boldsymbol{x}) = \frac{ p(\boldsymbol{x})-\min_{\boldsymbol{x} \in \mathcal{X}} p(\boldsymbol{x})}{ \max_{\boldsymbol{x} \in \mathcal{X}} p(\boldsymbol{x} )-\min_{\boldsymbol{x} \in \mathcal{X}} p(\boldsymbol{x}) }.
\end{align}


\subsection{Local Maximal Points and Gradient Flows}
In our method, density growing process aims to discover the local clusters and their boundary points, as well as dropping noise. Some similar definitions can be found in mode clustering \cite{chen2017statistical,dey2018graph} and persistence based clustering \cite{chazal2013persistence}. To illustrate the cluster growing process better, the definition of \textit{local maximal points} and \textit{gradient flows} proposed in Morse Theory \cite{chen2017statistical} are necessary.\\

\noindent\textbf{Definition 1.} (\textit{local maximal points} and \textit{gradient flows})
 Given density function $f(\boldsymbol{x})$, \emph{local maximal points} are $\mathcal{M}=\{ \boldsymbol{m}_i| \nabla f(\boldsymbol{m}_i)=0, |H(\boldsymbol{m}_i)|=|\nabla^2 f(\boldsymbol{m}_i)|<0 , \boldsymbol{m}_i \in \mathcal{X} \}$. Where $H(\boldsymbol{m}_i)$ is the Hessian matrix, and $|H(\boldsymbol{m}_i)| < 0$ means it is a negative definite matrix. For any point $\boldsymbol{x} \in \mathbb{R}^{N}$, there is a \textit{gradient flow} $\pi_{\boldsymbol{x}}:[0,1] \mapsto \mathbb{R}^{N}$, starting at $\pi_{\boldsymbol{x}}(0)=\boldsymbol{x}$ and ending in  $\pi_{\boldsymbol{x}}(1)=dest(\boldsymbol{x})$, where $dest(\boldsymbol{x}) \in \mathcal{M}$. The $i$-th \textit{local cluster} is the set of points converging to the same destination along \textit{gradient flow}, which is $v_i=\{\boldsymbol{x}|dest(\boldsymbol{x})=\boldsymbol{m}_i,\boldsymbol{m}_i \in \mathcal{M},\boldsymbol{x} \in \mathcal{X}\}$.\\

Based on continuous density obtained by LKDE, we are able to estimate gradient $\nabla f(\boldsymbol{x})$, the gradient flow $\pi_{\boldsymbol{x}}$, and the destination $dest(\boldsymbol{x})$. In practice, we just need to estimate these quantities in discrete sample sets, which reduces cost of computation considerably. Employing those concepts, the clustering centers are regarded as local maximal points and the process of searching for each point's cluster is regarded as following its gradient flow to its destination.

\section{Proposed Method}
\begin{figure*}[h]
	\centering
	    \includegraphics[width=6.6in]{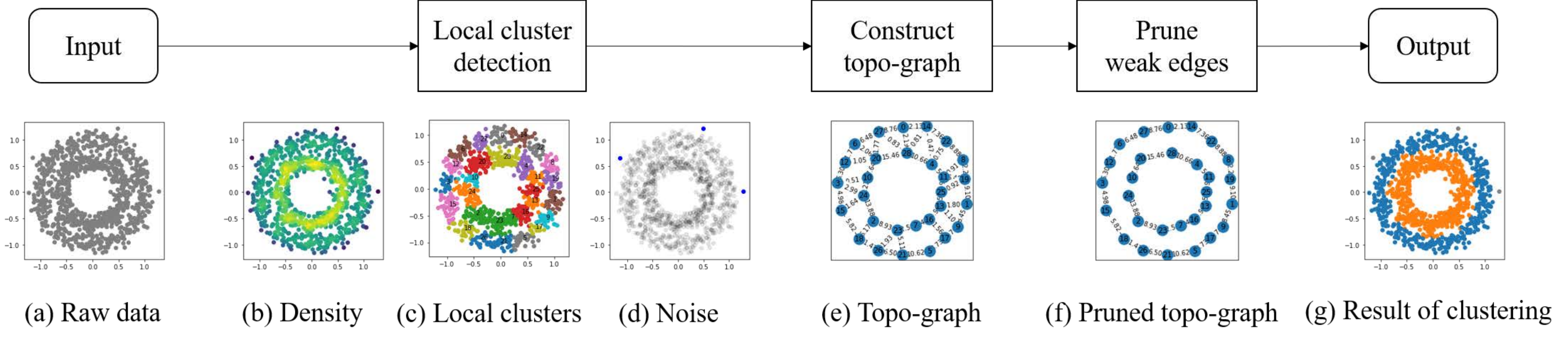}
	    \caption{An example to demonstrate the process of our GDT methods: after the density function in (b) has been estimated according to the raw data (two rings) in (a), local clusters and noise are detected through density growing process, which is showed by (c) and (d). (e) shows that each local cluster forms a vertices in the topological graph, and the connectivity between vertices are calculated. In (f), weak edges are pruned and two unconnected rings represent the two-ring structure. Finally, the clustering task is finished as shown in (g).}
	    \label{fig:exapmle of GDT using circle}
\end{figure*}

This section introduces the main modules of GDT, which can be summed up as three parts: local cluster detection in 3.1, topo-graph construction in 3.2 and edges pruning in 3.3.
Those three processes can be illustrated in Fig. \ref{fig:motivation_1d} and Fig. \ref{fig:exapmle of GDT using circle}.

\subsection{Local Cluster Detection}
To detect local clusters of $\mathcal{X}$ using density function $f(x)$, some sub-problems need to be solved: (1) how to estimate the real density function from $\mathcal{X}$ (corresponding to Fig. 3(b)); (2) how to find each local cluster formed by discrete data points efficiently (corresponding to Fig. 3(c)); (3) how to detect boundary points of two adjacent local clusters; (4) how to deal with noise(corresponding to Fig. 3(d)). This section develops according to these problems.

\paragraph{Density estimation.} 
LKDE is used for density estimation. For simplicity, we use the Gaussian kernels written as $\kappa(\cdot) = exp{(-\frac{(\cdot)^2}{2h_l^2})}$. According to Silverman's rule of thumb \cite{silverman1986density}, the optimal kernels' bandwidths are given by  $h_l = (\frac{4\hat \sigma_l^5}{3n})^{0.2}$, where $\hat \sigma_l$ is the standard deviation of the $l$-th dimension of the whole sample set $\{x_{1}^{(l)}, x_{2}^{(l)}, \cdots, x_{n}^{(l)}\}$.

\begin{algorithm}[h]
    \caption{Local cluster detection algorithm}\label{euclid1}
    \hspace*{\algorithmicindent} 
    \begin{algorithmic}[1]
        \Require sample set $\mathcal{X}$, neighborhood size for LKDE $k_d$ , neighborhood size for searching local clusters $k_s$, noise threshold $\epsilon$;
        \Ensure local clusters $V$, $\mathcal{B}$, $X'$;
        \State Initialization.
        \For{$i \leftarrow 0,n-1 $}
            \State $P[i]\gets f(\boldsymbol{x}_i)$ \Comment{using kd-tree $\mathcal{O}(k_d n\log{n})$}
            \State Calculate $\mathcal{N}_i$ \Comment{using kd-tree $\mathcal{O}(k_s n\log{n})$}
        \EndFor
        \State $idx \gets \arg$ sort$(-P)$  \Comment{Heapsort $\mathcal{O}(n\log{n})$}
        \State $\mathcal{B}=[\quad]$
        \While{$idx \neq \phi$}  
            \State $i \gets idx[0],\lambda \gets P[i]$
            \State $\mathcal{J} \gets \{j|j \in \mathcal{N}_i, P[j]>\lambda \}$ \Comment{ $\mathcal{O}(k_s)$}
            \For{$j \in \mathcal{J}$}
                \State $grad_{i\rightarrow j} = \frac{f(\boldsymbol{x}_j)-f(\boldsymbol{x}_i)}{d(\boldsymbol{x}_i,\boldsymbol{x}_j)} $, Eq. \ref{eq:Prx} \Comment{$\mathcal{O}(k_s)$}
            \EndFor
            \State $j \gets \arg\max_{j\in \mathcal{J}} grad_{i\rightarrow j}$
            \Comment{$\mathcal{O}(k_s)$}
            \If {$j \neq \phi$} \Comment{parent exists}
                \State $r_i \gets r_j$ 
                \If{$r_i \neq -1$}
                    \If{{$P[i]/P[r_i]<\epsilon$}}
                        \State $r_i \gets -1$ \Comment{drop noise}
                    \Else
                        \For{$s \in \mathcal{N}_i$} \Comment{save $\mathcal{B}$, $\mathcal{O}(k_s^2)$}
                            \If{$r_s \neq r_i$ and $i \in \mathcal{N}_s$}
                                \State $\mathcal{B}$.append($(i,s,r_i,r_s)$)
                            \EndIf
                        \EndFor
                    \EndIf
                \EndIf
            \EndIf
            
            \State idx.remove($i$) 
        \EndWhile
        \State $V=\{j:[\quad]$ for $j$ in set($R$)$\}$ \Comment{$\mathcal{O}(n)$}
        \For {$i$ in range(len($R$))}  \Comment{$\mathcal{O}(n)$}
            \State $V[r_i]$.append($i$) 
        \EndFor
    \end{algorithmic}
    Initialization: Extend $\mathcal{X}$ to $ X' \in \mathbb{R}^{n,d+4}$, where additional dimensions represent \textit{density}, \textit{index}, \textit{local clusters} and \textit{label}, all of which are initialized as samples' indexes. Denote $P=X'[:,-4],R=X'[:,-2], r_i=R[i]$ as density array, root array and $\boldsymbol{x}_i$'s root respectively.
\end{algorithm}

\begin{figure}[ht]
	\centering
		\subfigure[Density]{ 
			\includegraphics[width=0.15\linewidth]{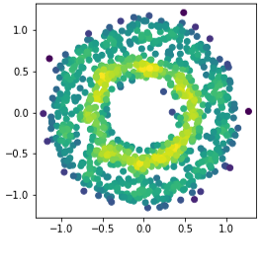}}
		\subfigure[$\lambda_{10}=0.9739$]{ 
			\includegraphics[width=0.15\linewidth]{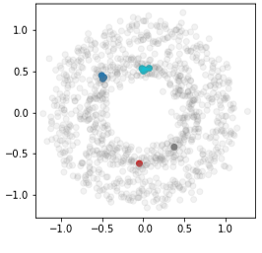}}
		\subfigure[$\lambda_{100}=0.9141$]{ 
			\includegraphics[width=0.15\linewidth]{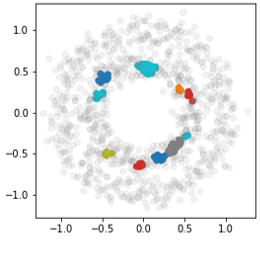}}\\
		\subfigure[$\lambda_{300}=0.7923$]{ 
			\includegraphics[width=0.15\linewidth]{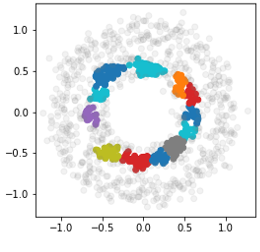}}
		\subfigure[$\lambda_{700}=0.5732$]{ 
			\includegraphics[width=0.15\linewidth]{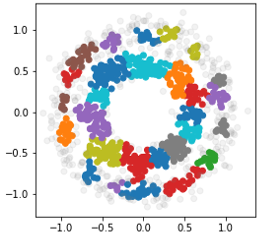}}
        \subfigure[$\lambda_{1000}=0$]{ 
			\includegraphics[width=0.15\linewidth]{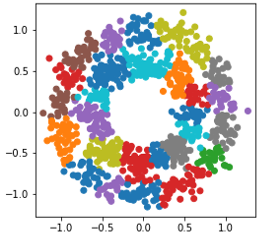}}
			\caption{An example of dynamic density growing process: (a) shows the density estimated by LKDE; and (b),(c),(d),(e),(f) shows the dynamic change of clustered set $Q_{\lambda_k}$ under different $\lambda_k$. The light grey points are not in the clustered set. The points with other different colors are elements born in clustered set, and different colors mean different clusters the points belong to. As $\lambda_k$ decays, $Q_{\lambda_k}$ grows, and when $k=1000$, it drops to 0, and every point has  into the clustered set.}
		\label{fig:density_growing_process}
\end{figure}

\paragraph{Density growing process.}
We introduce a density growing process where points with higher density birth earlier for local clusters detection. Fig. \ref{fig:density_growing_process} shows an instance.

Mathematically, density growing process can be illustrated with a series of super-level sets. The super-level set of $f(\boldsymbol{x})$ corresponding to level $\lambda$ is $L_{\lambda}^{+}=\{\boldsymbol{x}|\lambda\leq f(\boldsymbol{x})\}$. Given an descending-ordered series of level $\Lambda=\{\lambda_1,\lambda_2,\cdots,\lambda_t \}$, $\lambda_{k}>\lambda_{k+1}$ , $\mathcal{Q}_{\lambda_k} = \{(x_i,j)| x_i\in L_{\lambda_k}^{+}, 1\leq j\leq m\}$ is a clustered set with respect to $\lambda_k$, where $(\boldsymbol{x}_i,j)$ indicates that the sample point $\boldsymbol{x}_i$ belongs to $j$-th local cluster $v_j$. Note that $L_{\lambda_{k}}^{+} \subset L_{\lambda_{k+1}}^{+} $ and $\mathcal{Q}_{\lambda_{k}} \subset \mathcal{Q}_{\lambda_{k+1}}$. Therefore, as $\lambda$ descends from $\lambda_{k}$ to $\lambda_{k+1}$, $\mathcal{Q}_{\lambda_{k+1}} $ will be correspondingly calculated, which can be viewed as a process of new point $\boldsymbol{x}_{new}$ appearing and $\mathcal{Q}_{\lambda_{k}}$ growing 
into $\mathcal{Q}_{\lambda_{k+1}}$. We call that $\boldsymbol{x}_{new}$ is born at $\lambda_{k+1}$ if $ \boldsymbol{x}_{new} \in L_{\lambda_{k+1}}^{+}\setminus L_{\lambda_{k}}^{+}$. 

Specifically, in our case of density growing process, $\lambda$ gradually descends from 1 to 0. When the new point $\boldsymbol{x}_i$ births at $\lambda$, to calculate $Q_{\lambda}$, we need to decide which cluster it belongs to. Employing the concepts of \textit{local maximal points} and \textit{gradient flow}, we regard $\boldsymbol{x}_i$ which is the center point forming a new local cluster in case (1) as a \textit{local maximal point}, and our target turns to how to identify it. Case (2) where $\boldsymbol{x}_i$ belongs to the existing local cluster is regarded as searching for the parent point of $\boldsymbol{x}_i$ along the \textit{gradient flow}. Therefore, we identify which local cluster each point belongs to according to the following rules, where local clusters are equivalent to \textit{Morse-Smale complexes} in Morse Theory:

    (1) If $\nabla f(\boldsymbol{x}_i)=0$ and $|H(\boldsymbol{x}_i)|<0$, $\boldsymbol{x}_i$ is a center of a local cluster, and $\boldsymbol{x}_i \in \mathcal{M}$;
    
    (2) If $\nabla f(\boldsymbol{x}_i) \neq0$ and $\boldsymbol{x}_j-\boldsymbol{x}_i$ is the gradient direction from $\boldsymbol{x}_i$, $\boldsymbol{x}_j \in \mathcal{X}$ is the parent point of $\boldsymbol{x}_i$ along the gradient flow. $\boldsymbol{x}_i$'s parent is also denoted as $Pr(\boldsymbol{x}_i)$, sharing the same label with $\boldsymbol{x}_i$.



 For finding the root (or destination) of $\boldsymbol{x}_i$ denoted by $Rt(\boldsymbol{x}_i)$ along the \textit{gradient flow}, we estimate the gradient direction around $\boldsymbol{x}_i$ by discrete maximum directional derivative. When $\boldsymbol{x}_i$ is born at $\lambda_k$, $\boldsymbol{x}_i$'s parent node $Pr(\boldsymbol{x}_i)$ is defined as one of $\boldsymbol{x}_i$'s neighbors born before $\boldsymbol{x}_i$, who has maximum directional derivative starting from $\boldsymbol{x}_i$ and ending in $Pr(\boldsymbol{x}_i)$. That is, for $\boldsymbol{x}_i \in L_{\lambda_{k}}^{+}\setminus L_{\lambda_{k-1}}^{+}$,
 
\begin{align}
    \label{eq:Prx}
    Pr(\boldsymbol{x_i})={\arg\max}_{\boldsymbol{x}_p\in L_{\lambda_{i-1}}^{+} \cap \mathcal{N}_{\boldsymbol{x}_i}}\frac{f(\boldsymbol{x}_p)-f(\boldsymbol{x}_i)}{d(\boldsymbol{x}_i,\boldsymbol{x}_p)},
\end{align}
where $\mathcal{N}_i$ is the $k_s$ nearest neighborhood system of $\boldsymbol{x}_i$, and $|\mathcal{N}_i|=k_s$. Using the Eq. \ref{eq:Prx}, we can determine each point's parent as well as its label. 
In case (1), $\boldsymbol{x}_i$ is a local maximal point of $f(\boldsymbol{x}_i)$,  indicating that $Pr(\boldsymbol{x}_i)=\boldsymbol{x}_i$, and $\boldsymbol{x}_i$ belongs to a new cluster differing from all existing clusters;
In case (2), after the $Pr(\boldsymbol{x}_i)$ is identified, and $Pr(\boldsymbol{x}_i) \neq \boldsymbol{x}_i$, $\boldsymbol{x}_i$ inherits the label of $Pr(\boldsymbol{x}_i)$.

To sum up, in density growing process, the following conclusions hold true:
    
    (1) $\boldsymbol{x}_i \in \mathcal{M} \iff Pr(\boldsymbol{x}_i)=\boldsymbol{x}_i = \boldsymbol{m}_j \in \mathcal{M} $;
    
    (2) $\boldsymbol{x}_i \in v_j \iff Rt(\boldsymbol{x}_i)=Pr(\cdots Pr(\boldsymbol{x}_i))=\boldsymbol{m}_j$.

\paragraph{Boundary points.}
Define the boundary pair set of $v_i$ and $v_j$ as $\mathcal{B}=\{ (\boldsymbol{x}_p,\boldsymbol{x}_q) |\boldsymbol{x}_p \in \mathcal{N}_{\boldsymbol{x}_q}, \boldsymbol{x}_q \in \mathcal{N}_{\boldsymbol{x}_p}\}$, where $\mathcal{N}_{\boldsymbol{x}_p}$ is the neighborhood system of $\boldsymbol{x}_p$, $|\mathcal{N}_{\boldsymbol{x}_p}|=|\mathcal{N}_{\boldsymbol{x}_q}|=k_s$. $\mathcal{B}$ can be efficiently detected along with density growing process, without much more computation. $\mathcal{B}$ is helpful to calculate the connectivity between two local clusters later. 

\paragraph{Noise dropping.}
In noisy case, $\boldsymbol{x}_i \in v_j$ is identified as noise if $ |\frac{f(\boldsymbol{x})}{\max_{\boldsymbol{x} \in v_j }f(\boldsymbol{x})}|<\epsilon$ for a given $\epsilon$.
\\

\noindent We offer the Python-styled pseudo-code of \textbf{Algorithm 1} for local cluster and boundary point detection with the note on time complexity analysis.

\subsection{Topo-graph Construction}
As local clusters $ (v_j)_{1\leq j \leq m}$ are obtained, we can construct a topological graph, \textit{graph of density topology},  for revealing the relationships between local clusters based on their connectivity(corresponding to Fig. 3(e)). To define the connectivity between $v_i$ and $v_j$ written as $e_{i,j}$, the boundary pair set $\mathcal{B}$ will be used.
The connectivity of $v_i$ and $v_j$ is $e_{i,j}=w_{i,j}\cdot \gamma_{i,j}$, derived from two aspects: 

(1) The summation of density of mid points in pairs: $\frac{\boldsymbol{x}_p + \boldsymbol{x}_q}{2}$, where $(\boldsymbol{x}_p,\boldsymbol{x}_q) \in \mathcal{B}$. Intuitively, the more points in boundary pair set and the higher the density of middle points of the boundary pair, the stronger the connectivity of two local clusters should be. 
\begin{gather}
\label{eq:connectivity1}
  w_{i,j}= \sum_{ (\boldsymbol{x}_p,\boldsymbol{x}_q) \in \mathcal{B} }{   f(\frac{\boldsymbol{x}_p+\boldsymbol{x}_q}{2}) }.
\end{gather}

(2) The difference of density between peaks of $v_i$ and $v_j$ as a modifying term for connectivity. Assert that similar local clusters have close density. 

Based on the two aspects, connectivity is defined as
\begin{gather}
\label{eq:connectivity2}
  \gamma_{i,j}=\min\{ \frac{f(\boldsymbol{m}_i)}{f(\boldsymbol{m}_j)} , \frac{f(\boldsymbol{m}_j)}{f(\boldsymbol{m}_i)} \}.
\end{gather}
In practice, we find it better to add a transformation function, and the Eq. \ref{eq:connectivity1} and Eq. \ref{eq:connectivity2} can be written as
\begin{equation}
\label{eq:connectivity_plus}
\begin{cases}
    &  w_{i,j}= \sum_{ (\boldsymbol{x}_p,\boldsymbol{x}_q) \in \mathcal{B}}{   \Phi_1 ( f(\frac{\boldsymbol{x}_p+\boldsymbol{x}_q}{2}) });\\
    &  \gamma_{i,j}=\Phi_2 (\min\{ \frac{f(\boldsymbol{m}_i)}{f(\boldsymbol{m}_j)} , \frac{f(\boldsymbol{m}_j)}{f(\boldsymbol{m}_i)} \}),
\end{cases}
\end{equation}
where $\Phi_1$ and $\Phi_2$ are the transform functions, which are monotonically increasing and non-negative in $[0, 1]$. Specifically, we choose $\Phi_1(x)=\Phi_2(x)=x^2$ to magnify the differences.

\begin{figure}[ht]
	\centering
		\subfigure[Local clusters]{ 
			\includegraphics[width=0.2\linewidth]{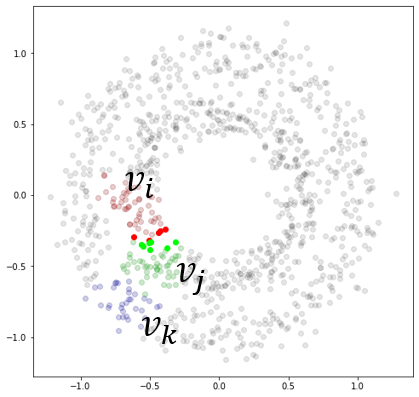}}
		\subfigure[Boundary pairs]{ 
			\includegraphics[width=0.2\linewidth]{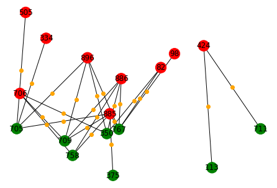}}
		\subfigure[Sketch density]{ 
			\includegraphics[width=0.2\linewidth]{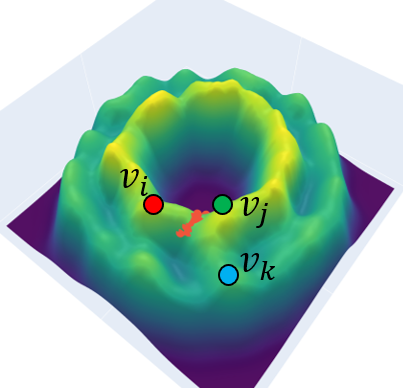}}
\caption{An illustration of boundary points: In (a), the green, red and blue points belong to $v_i$, $v_j$ and $v_k$, represented by vertices $v_i, v_j, v_k$ respectively. In (b), boundary pairs crossing $v_i$ and $v_j$ are shown with full lines, and the mid points of pairs are colored with orange, with their corresponding density shown in (c). }
		\label{fig:boundary points}
\end{figure}


\begin{algorithm}[H]
    \caption{Topo-graph construction and pruning}\label{euclid2}
    
    \begin{algorithmic}[1]
    \Require $X'$, boundary pair set $\mathcal{B}$, LKDE  $f(\cdot)$, $\alpha$
    \Ensure $E$
        \State initial $E \gets \{\},\tilde{E} \gets \{\},\gamma \gets \{\}$, denote $Rt(\boldsymbol{x}_i)$ as $r_i$
        \For{$(i,j,r_i,r_j) \in \mathcal{B}$} \Comment{$\mathcal{O}(|\mathcal{B}|)<\mathcal{O}(n)$}
            \State $E_{r_i,r_j}, \tilde{E}_{r_i}, \tilde{E}_{r_j},\gamma_{r_i,r_j}=$ None, None, None, None.
        \EndFor
        \For{$(i,j,r_i,r_j) \in \mathcal{B}$} \Comment{$\mathcal{O}(|\mathcal{B}|)<\mathcal{O}(n)$}
            \If{$\gamma_{r_i,r_j}$ is None }
                \State Calculate $\gamma_{r_i,r_j} $ according to Eq. \ref{eq:connectivity_plus}
                \State $\gamma_{r_i,r_j} \gets \gamma_{r_j,r_i}$
            \EndIf
            \State Calculate $w_{r_i,r_j}$ according to Eq. \ref{eq:connectivity_plus}
            \State $s \gets \gamma_{r_i,r_j}w_{r_i,r_j}$  \Comment{$\mathcal{O}(k_d \log{n})$}
            \State $E_{r_i,r_j}+=s$
            \State $E_{r_j,r_i} \gets E_{r_i,r_j}$
            \State $\tilde{E}_{r_i} \gets E_{r_i,r_j}$, if $E_{r_i,r_j}>\tilde{E}_{r_i}$
            \State $\tilde{E}_{r_j} \gets E_{r_i,r_j}$, if $E_{r_i,r_j}>\tilde{E}_{r_j}$
        \EndFor
        
        \For{$E_{i,j} \in E$} \Comment{$\mathcal{O}(|E|)<\mathcal{O}(|\mathcal{B}|)$}
            \If{$\frac{E_{i,j}}{\tilde{E}_i}<\alpha$ or $\frac{E_{i,j}}{\tilde{E}_j}<\alpha$}
                \State $E_{i,j} \gets 0$ 
            \EndIf
        \EndFor
    \end{algorithmic}
\end{algorithm}

\subsection{Topo-graph Pruning}
This section introduces how to prune the weak edges of $G= (V,E)$ while retaining strong edges to get more reliable topology structure(corresponding to Fig. 3(f)).\\ 

Denote the strongest edge of $v_i$ as $\tilde{e}_i=\max_{j} e_{i,j}$, the relative value of $e_{i,j}$ as $r_{i,j}=\frac{e_{i,j}}{\tilde{e}_i} \in [0,1]$. Use $\mathbbm{1}_{i,j} \in \{0,1\}$ to identify whether $e_{i,j}$ exists or not after pruning: If $\mathbbm{1}_{i,j}=0$, $e_{i,j}$ will be cut, and vice versa. The objective for optimization for each $i$ is

\begin{equation*}
\begin{aligned}
   &  \min_{(\mathbbm{1}_{i,j})_{1\leq j\leq m}}
   && \mathcal \sum_{j=1}^m {\mathbbm{1}_{i,j}(r'_{i,j}-1)^2+\beta (1-\mathbbm{1}_{i,j}) (r'_{i,j}-r_{i,j})^2};\\
   & s.t. 
   && \tilde{e}'_i=\tilde{e}_i;\quad e'_{i,j}= \mathbbm{1}_{i,j} \cdot e_{i,j} ;\quad r'_{i,j} = e'_{i,j}/e'_i\\
\end{aligned}
\end{equation*}
where the first term aims to cut off weak edges less than $1$, which is $\max_{j}(r_{i,j})$, the second term aims to preserve strong edges, and $\beta$ is a weight for balance. For the optimized value of $\mathbbm{1}_{i,j}$, two cases need to be considered: When $e_{i,j}$ is cut, $\mathcal{L}_i \vert_{\mathbbm{1}_{i,j}=0}=\beta r_{i,j}^2$; Otherwise, $\mathcal{L}_i \vert_{\mathbbm{1}_{i,j}=1}=\beta (r_{i,j}-1)^2$. When $\mathcal{L}_i \vert_{\mathbbm{1}_{i,j}=0}<\mathcal{L}_i \vert_{\mathbbm{1}_{i,j}=1}$, $e_{i,j}$ shall be cut, and the solution is $r_{i,j}<\frac{1}{\sqrt{\beta}+1}$.To summary,

\begin{equation*}
\begin{cases}
    &  \mathbbm{1}_{i,j}=0\quad\quad r_{i,j}<\frac{1}{\sqrt{\beta}+1}\\
    &  \mathbbm{1}_{i,j}=1 \quad\quad else
\end{cases}
\end{equation*}

Once $\alpha=\frac{1}{\sqrt{\beta}+1} \in (0,1]$ is given, $(\mathbbm{1}_{i,j})_{1\leq j \leq m}$ can be determined. However, the process of cutting edges may not be symmetric, that is $\mathbbm{1}_{i,j} \neq \mathbbm{1}_{j,i}$. The greedy strategy is employed to cut weak edges as much as possible: once it satisfies that $\mathbbm{1}_{i,j}=0$ or $\mathbbm{1}_{j,i} = 0$ , the edge $e_{i,j}$ will be cut.

\subsection{Properties of the method}
\paragraph{Computational complexity.}
In the analysis, given $n$ samples, we assume that density based algorithms work on low dimensional case $(d\ll n)$, and thus the dimension $d$ can be viewed as a constant. Besides, $k_d$ and $k_s$ are manually specified constants. The time complexity is just correlated with the number of samples $n$. We can reach the total time complexity of $\mathcal{O}(n\log{n})$. For more details, see Supplementary A.1.
Compared with k-means with $\mathcal{O}(cnt)$ complexity \cite{sklearn_api}, where $c$ and $t$ are the numbers of clusters and iterations, spectral clustering with $\mathcal{O}(n^3)$, mean-shift with $\mathcal{O}(n\log{n})$, OPTICS with $\mathcal{O}(n\log{n})$, DBSCAN with $\mathcal{O}(n^2)$ and accelerated HDBSCAN with the complexity of $\mathcal{O}(n\log{n})$, GDT is competitive.


\paragraph{Density growing process.}
HDBSCAN adopts the 'backward strategy', dropping the small point set as noises or separating large point set as a new cluster according to the minimum cluster size and \textit{relative excess of mass} for the cluster tree, leading to excessive sample loss. In contrast, our method adopts the 'forward strategy', as the local clusters accept the near points to grow according to the approximated gradient flows. Noise will be dropped if the relative density is smaller than the given threshold $\epsilon$, allowing a more steerable noise dropping, and the experiments show that the strategy is more stable in avoiding the excessive loss of sample points.

\paragraph{Topo-graph construction.}
Our method is able to construct the topological graph to describe the connectivity between local clusters. The pruned cluster trees established by HDBSCAN can also be viewed as a tree structured graph for evaluating connectivity between points rather than local clusters, which can't handle bridge-like noise (seeing Fig. \ref{fig:comparision}) between two clusters. However, other density-based algorithms like DBSCAN and mean-shift is not able.  Besides, the defined connectivity takes both boundary points and difference of local clusters into consideration, which is a more direct reflection of the relationship between local clusters. And our experiments prove it an appropriate definition that can correctly reveal the number of class and establish topological structure without any prior knowledge.

\section{Experiments}
GDSFC is evaluated on both classification and segmentation tasks, with other clustering algorithms compared. The hyper-parameters used in the experiments and the analysis of them is attached in Supplementary A.2 and A.3.
\subsection{Classification}
We evaluate our method on 5 toy datasets and 5 real-world datasets on the classification tasks.
\begin{table}[h]
\centering
\caption{The description of real-world datasets for evaluation }
    \begin{tabular}{cccccc}
    \toprule
                        & Iris    & Wine   & Glass     & Hepatitis        & Cancer  \\
    \midrule
    classes             & 2       & 3    & 6         & 2                & 2       \\
    sample              & 150       & 178    & 214        & 154                & 569      \\
    dimension           & 4       & 13    & 9         & 19                & 30       \\
    discrete            & 0       & 0    & 0        & 13               & 0       \\
    continuous          & 4       & 12    & 9        & 6                & 30      \\
    \bottomrule
    \end{tabular}
\label{Table:dataset}
\end{table}
\paragraph{Datasets.} 10 individual datasets are used to evaluation, 5 of which are real-world datasets \cite{FisherIris,AeberhardWine,IanGlass,Diaconis1983Hepatitis,StreetBreastCancer}, representing a large variety of application domains and data characteristics. The information on them is listed in Table. \ref{Table:dataset}, and the missing values are filled with mean. In addition, in the toy dataset 'Circles' and 'Moons', we manually add a Gaussian noise to each point, with zero means and standard deviation $\sigma_{moons} = 0.15$ and $\sigma_{circles} = 0.1$, which is very high noise levels for increasing the difficulty for clustering tasks.
\paragraph{Algorithms.}
Our method, denoted by 'GDT', is compared with density based methods: (1)Hierarchical Density-Based Spatial Clustering of Applications with Noise, denoted by 'HDBSCAN', (2)Mean-shift and (3)Ordering Points to Identify the Clustering Structure, denoted by 'OPTICS'.  Besides, some other unsupervised learning methods are also compared, including (1)Spectral Clustering, denoted by 'Spectral' and (2) k-means.

\paragraph{Measures.}
The measures reported are Accuracy, F-score \cite{Larsen1999}, and Adjusted Rand Index \cite{Hubert1985}, which is denoted by 'Acc', 'FScore' and 'ARI' respectively. Accuracy is the ratio of true label to sample number, ranging from $0$ to $1$, and the closer it is to $1$, the better the result is. F-score is the index evaluating both each class's accuracy and the bias of the model, ranging from $0$ to $1$. Adjusted Rand Index is a measure of agreement between partitions, ranging from $-1$ to $1$, and if it is less than $0$, the model does not work in the task. In addition, because density-based algorithm drops some data points as noise, we also report the fraction of samples assigned to clusters, denoted by '$\%$covered'. Spectral clustering and k-means are not able to drop noise, so we have not taken their comparison of '$\%$covered' into account.
\begin{figure}[h]
	\centering
	    \includegraphics[width=4.5in]{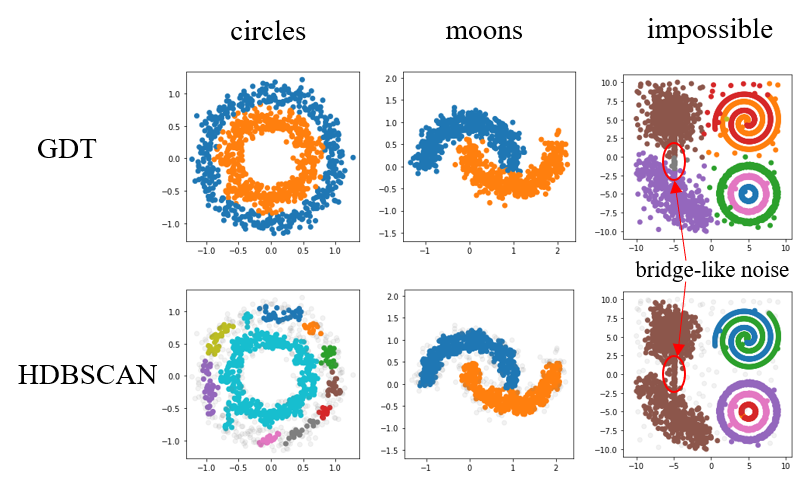}
		\caption{Visualization on two of 2-d toy datasets at high noise levels. Points with different colors represent different clustering labels, and the light grey points are the noise points inferred by algorithms. While GDT and HDBSCAN are the top two algorithms according to the comprehensive quality measures, HDBSCAN drop too many points as noise. In contrast, GDT covers all the points, and demonstrates excellent performance, indicating GDT's competence of dealing with highly-noisy datasets. Note that in the visualization of 'Impossible' dataset, for HDBSCAN, a single bridge-like noise point dramatically increases the connectivity between two clusters, misleading the two weakly connected clusters sharing one label, while GDT is more robust to it due to its connectivity defined on local clusters rather than individual points.  }
		\label{fig:comparision}
\end{figure}

\begin{table*}[h]
\caption{ Results of experiments }
\resizebox{\textwidth}{!}{
    \begin{tabular}{cccccccccccc}
    \toprule
    \textbf{}                   &             & Circles & Moons  & Impossible & S-set   & Smile  & Iris   & Wine    & Cancer & Glass  & Hepatitis       \\
    \midrule
    \multirow{4}{*}{DGSFC}      & Fscore          & \textbf{0.9570} & 0.9850 & \textbf{0.9994} & 0.9988 & \textbf{1.0000} & \textbf{0.9397} & \textbf{0.8159} & \textbf{0.8648} & \textbf{0.5759} & \textbf{0.7322} \\
                                & ARI         & \textbf{0.8352} & 0.9408 & \textbf{0.9990} & 0.9974 & \textbf{1.0000} & \textbf{0.8345} & \textbf{0.5532} & \textbf{0.6103} & 0.2147          & \textbf{0.3958} \\
                                & ACC         & \textbf{0.9570} & 0.9850 & \textbf{0.9992} & 0.9988 & \textbf{1.0000} & \textbf{0.9400} & \textbf{0.8202} & 0.8295          & 0.5127          & 0.7468          \\
                                & \%cover & \textbf{1.0000} & \textbf{1.0000} & \textbf{1.0000} & \textbf{1.0000} & \textbf{1.0000} & \textbf{1.0000} & \textbf{1.0000} & \textbf{1.0000} & 0.7383          & \textbf{1.0000} \\
    \hline
    \multirow{4}{*}{HDBSCAN}    & Fscore          & 0.7387          & \textbf{0.9919}          & 0.8235          & 0.9987          & \textbf{1.0000} & 0.5715          & 0.5435          & 0.7848          & 0.5083          & 0.7073          \\
                                & ARI         & 0.8162          & \textbf{0.9678}          & 0.8010          & 0.9973          & \textbf{1.0000} & 0.5759          & 0.3034          & 0.4041          & 0.2373          & 0.0506          \\
                                & ACC         & 0.7117          & \textbf{0.9919}          & 0.8713          & 0.9988          & \textbf{1.0000} & 0.6803          & 0.6353          & 0.8160          & \textbf{0.5789} & \textbf{0.7655} \\
                                & \%cover & 0.6590          & 0.8630          & 0.9961          & 0.9608          & \textbf{1.0000} & 0.9800          & 0.9551          & 0.8120          & 0.7103          & 0.9416          \\
    \hline
    \multirow{4}{*}{mean-shift} & Fscore          & 0.3070          & 0.4319          & 0.5694          & 0.4502          & 0.7347          & 0.7483          & 0.4613          & 0.8569          & 0.3812          & 0.6625          \\
                                & ARI         & -0.0026         & 0.0711          & 0.6482          & 0.6148          & 0.7078          & 0.5613          & 0.1650          & 0.5595          & 0.2954          & 0.0807          \\
                                & ACC         & 0.2206          & 0.2786          & 0.6771          & 0.5857          & 0.8170          & 0.6552          & 0.3595          & \textbf{0.8558}          & 0.4731          & 0.6240          \\
                                & \%cover & 0.9110          & 0.8470          & 0.9700          & 0.8724          & 0.9180          & 0.7733          & 0.8596          & 0.9262          & \textbf{0.8692} & 0.8117          \\
    \hline
    \multirow{4}{*}{OPTICS}     & Fscore          & 0.3533          & 0.5412          & 0.8110          & \textbf{0.9996}          & 0.9594          & 0.4489          & 0.5223          & 0.4413          & 0.4040          & 0.1041          \\
                                & ARI         & 0.0467          & 0.1321          & 0.7903          & \textbf{0.9991}          & 0.9018          & 0.1193          & 0.1446          & 0.1063          & \textbf{0.3344} & -0.0175         \\
                                & ACC         & 0.2158          & 0.3730          & 0.8649          & \textbf{0.9996}          & 0.9272          & 0.3065          & 0.4024          & 0.2897          & 0.4656          & 0.0566          \\
                                & \%cover & 0.4310          & 0.4290          & 0.9227          & 0.8960          & 0.6320          & 0.4133          & 0.4607          & 0.3761          & 0.6121          & 0.6883          \\
    \hline
    \multirow{3}{*}{spectral}   & Fscore          & 0.5079          & 0.7720          & 0.5588          & 0.0416          & 0.6755          & 0.8988          & 0.3287          & 0.4838          & 0.3843          & 0.5648          \\
                                & ARI         & -0.0007         & 0.2952          & 0.6324          & -0.0001         & 0.5524          & 0.7437          & -0.0009         & 0.0000          & 0.2082          & -0.0042         \\
                                & ACC         & 0.5080          & 0.7720          & 0.5944          & 0.0808          & 0.7030          & 0.9000          & 0.3596          & 0.6274          & 0.4860          & 0.5195          \\
    \hline
    \multirow{3}{*}{k-means}    & Fscore          & 0.5018          & 0.7579          & 0.4819          & 0.9976          & 0.6656          & 0.8918          & 0.7148          & 0.8443          & 0.5073          & 0.7050          \\
                                & ARI         & -0.0010         & 0.2655          & 0.6218          & 0.9950          & 0.5468          & 0.7302          & 0.3711          & 0.4914          & 0.2716          & 0.0191          \\
                                & ACC         & 0.5020          & 0.7580          & 0.5191          & 0.9976          & 0.6960          & 0.8933          & 0.7022          & 0.8541          & 0.5421          & 0.7403 \\
    \bottomrule
    \end{tabular}
}
\label{Table:Results}
\end{table*}

\textbf{Results.} \quad Results obtained in our experiments are shown in Table. \ref{Table:Results}, with highest values highlighted in bold. It demonstrates that GDT outperforms the other methods in a large majority of the datasets. In the datasets of 'Moons', 'S-set' and 'Glass', GDT does not perform best, but its 'Fscore', 'ARI' and 'ACC' are very close to the highest measures, with the highest cover rate in 'Moons' and 'S-set, showing it is more practical for application, while other density-based algorithms tend to drop excessive points for exchanging for the good performance in precision.
In addition, visualization on certain 2-d toy datasets are compared with the algorithm ranking second in Fig. \ref{fig:comparision}, which indicates that even in the very noisy case, 
GDT can also distinguish the clusters effectively, and drop only a small percent of sample points. Other visualization comparisons are shown in Supplementary. A4.

\subsection{Segmentation}
\begin{figure}[H]
	\centering
	    \includegraphics[width=4.5in]{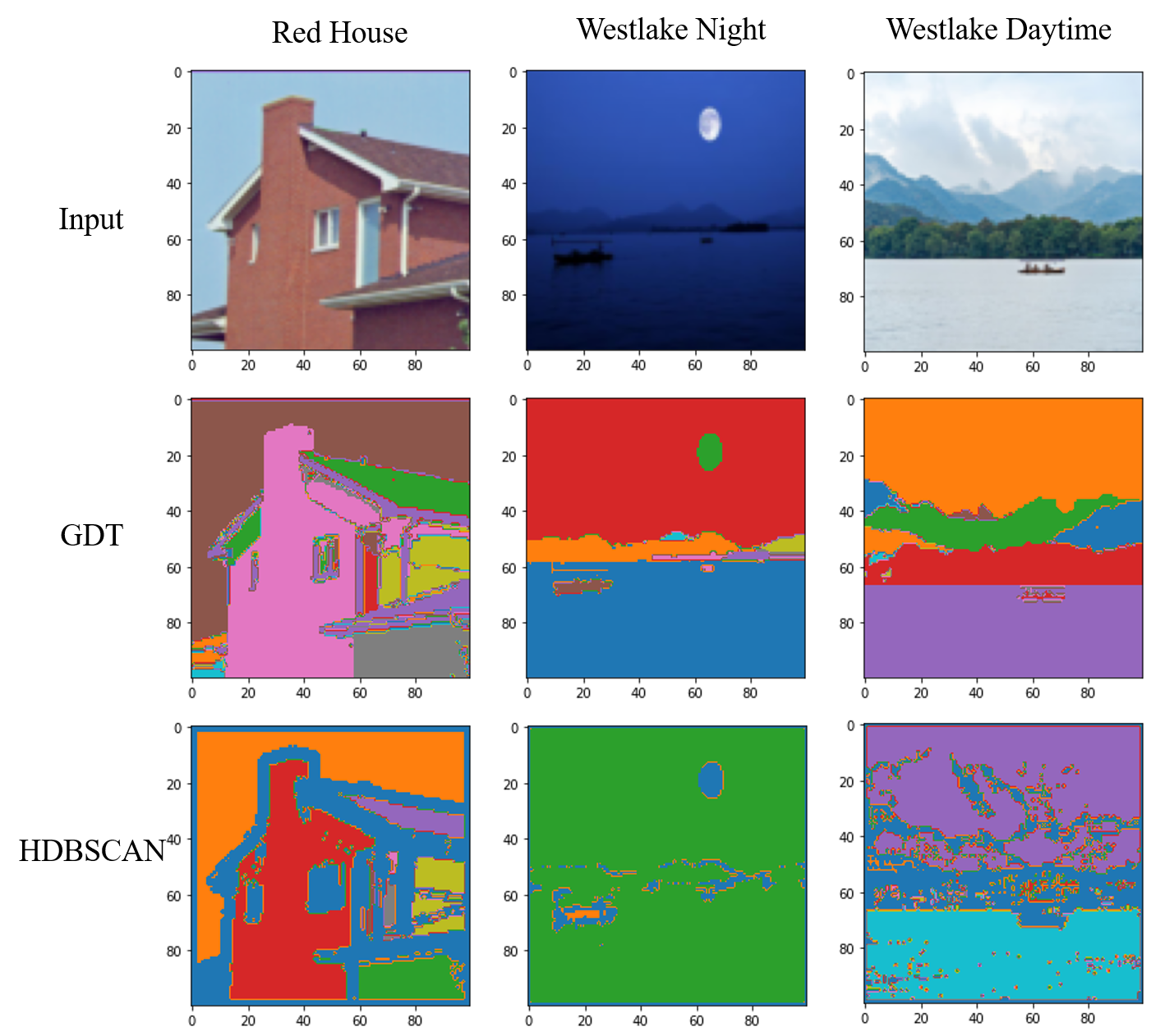}
		\caption{Image segmentation results: The first row is the raw images, and the second and the third is the segmentation results obtained by GDT and HDBSCAN. By visualization, our method outperforms HDBSCAN by a large margin. }
		\label{fig:segmentation}
\end{figure}
As a clustering method, we also evaluate GDT on image data for unsupervised segmentation. In the segmentation task, the input is an image, with each pixel a 5-d sample point: $(r, g, b, x, y)$, where $r, g, b$ represent 3 color channels of red, green and blue respectively, and $x, y$ represent the location of the pixel in the image. In our experiment, an image region is defined by all the pixels associated with the same local clusters in the joint domain. And the pruned graph allows the image region to share the same label, thus forming a bigger region. Besides, segmentation task done by HDBSCAN is also showed for comparison. The excessive sample loss of HDBSCAN leads to the limitation in practical application, whereas GDT can cover almost all the samples. The results on simple imagine is showed in Fig. \ref{fig:segmentation} for segmentation. And the other segmentation results are attached to Supplementary A.5.

\section{Conclusion}
A novel density-based clustering approach has been proposed in our paper. It provides: (1) local clusters detection algorithm, which is based on density growing process, functioning with boundary points discovery and noise dropping as well. (2) graph of density topology establishment, constructing the topological graph for evaluating the connectivity between clusters and pruning the weak edges for getting a more stable structure for label sharing. Our experimental evaluation has demonstrated that our method outperforms significantly better and more stable than other state-of-the-art methods on a wide variety of datasets. In the future work, we will extend our work to integration of semi-supervision and deep neural networks as well as more complete analysis on theoretical mechanism. Besides, emphasis will also be taken on hyper-parameter tuning and reduction.

\newpage
\bibliographystyle{plain} 
\bibliography{reference.bib}


\clearpage

\renewcommand{\algorithmicrequire}{\textbf{Input:}}  
\renewcommand{\algorithmicensure}{\textbf{Output:}} 





\centerline{\Large\bf Supplementary}
\setcounter{equation}{0}
\setcounter{figure}{0}
\setcounter{table}{0}

\textbf{Note}: The abbreviation of our method is GDT(graph of density topology) or DGSFC(density growing based structure finding and clustering)

\section*{A.1 Analysis of time complexity}
\setcounter{algorithm}{0}
\begin{algorithm}[H]
    \caption{Local cluster detection algorithm}\label{appeuclid1}
    \hspace*{\algorithmicindent} 
    \begin{algorithmic}[1]
        \Require sample set $\mathcal{X}$, neighborhood size for LKDE $k_d$ , neighborhood size for searching local clusters $k_s$, noise threshold $\epsilon$;
        \Ensure local clusters $V$, $\mathcal{B}$, $X'$;
        \State Initialization.
        \For{$i \leftarrow 0,n-1 $}
            \State $P[i]\gets f(\boldsymbol{x}_i)$ \Comment{using kd-tree $\mathcal{O}(k_d n\log{n})$}
            \State Calculate $\mathcal{N}_i$ \Comment{using kd-tree $\mathcal{O}(k_s n\log{n})$}
        \EndFor
        \State $idx \gets \arg$ sort$(-P)$  \Comment{Heapsort $\mathcal{O}(n\log{n})$}
        \State $\mathcal{B}=[\quad]$
        \While{$idx \neq \phi$}  
            \State $i \gets idx[0],\lambda \gets P[i]$
            \State $\mathcal{J} \gets \{j|j \in \mathcal{N}_i, P[j]>\lambda \}$ \Comment{ $\mathcal{O}(k_s)$}
            \For{$j \in \mathcal{J}$}
                \State $grad_{i\rightarrow j} = \frac{f(\boldsymbol{x}_j)-f(\boldsymbol{x}_i)}{d(\boldsymbol{x}_i,\boldsymbol{x}_j)} $, Eq. 3 \Comment{$\mathcal{O}(k_s)$}
            \EndFor
            \State $j \gets \arg\max_{j\in \mathcal{J}} grad_{i\rightarrow j}$
            \Comment{$\mathcal{O}(k_s)$}
            \If {$j \neq \phi$} \Comment{parent exists}
                \State $r_i \gets r_j$ 
                \If{$r_i \neq -1$}
                    \If{{$P[i]/P[r_i]<\epsilon$}}
                        \State $r_i \gets -1$ \Comment{drop noise}
                    \Else
                        \For{$s \in \mathcal{N}_i$} \Comment{save $\mathcal{B}$, $\mathcal{O}(k_s^2)$}
                            \If{$r_s \neq r_i$ and $i \in \mathcal{N}_s$}
                                \State $\mathcal{B}$.append($(i,s,r_i,r_s)$)
                            \EndIf
                        \EndFor
                    \EndIf
                \EndIf
            \EndIf
            
            \State idx.remove($i$) 
        \EndWhile
        \State $V=\{j:[\quad]$ for $j$ in set($R$)$\}$ \Comment{$\mathcal{O}(n)$}
        \For {$i$ in range(len($R$))}  \Comment{$\mathcal{O}(n)$}
            \State $V[r_i]$.append($i$) 
        \EndFor
    \end{algorithmic}
    Initialization: Extend $\mathcal{X}$ to $ X' \in \mathbb{R}^{n,d+4}$, where additional dimensions represent \textit{density}, \textit{index}, \textit{local clusters} and \textit{label}, all of which are initialized as samples' indexes. Denote $P=X'[:,-4],R=X'[:,-2], r_i=R[i]$ as density array, root array and $\boldsymbol{x}_i$'s root respectively.
\end{algorithm}

Because $k_d, k_s$ and $d$ can be viewed as constant, we ignore them during the analysis of time complexity.

For Algorithm 1, when using LKDE $f(\boldsymbol{x}_i)$ to estimate the density or searching the neighbor hood $\mathcal{N}_i$, $k_d$ or $k_s$ neighbors need to be found, which consumes $\mathcal{O}({k_d \log{n}})$ and  $\mathcal{O}(k_s \log{n})$ respectively by using k-d tree. Consider there are total $n$ points and constructing k-d tree costs $\mathcal{O}(n\log{n})$, the total time complexity from line 2 to line 4 is $\mathcal{O}(n\log{n})$. The following sort operation costs $\mathcal{O}(n\log{n})$. The main loop procedure repeats $n$ times, and within each loop, the time complexity is a constant, so its time complexity is $\mathcal{O}(n)$. The rest parts of Algorithm 1 cost $\mathcal{O}(n\log{n})$. In summary, Algorithm 1 costs $\mathcal{O}(n\log{n})$.

\begin{algorithm}[H]
    \caption{Topo-graph construction and pruning}\label{euclid2app}
    
    \begin{algorithmic}[1]
    \Require $X'$, boundary pair set $\mathcal{B}$, LKDE  $f(\cdot)$, $\alpha$
    \Ensure $E$
        \State initial $E \gets \{\},\tilde{E} \gets \{\},\gamma \gets \{\}$, denote $Rt(\boldsymbol{x}_i)$ as $r_i$
        \For{$(i,j,r_i,r_j) \in \mathcal{B}$} \Comment{$\mathcal{O}(|\mathcal{B}|) \leq \mathcal{O}(n)$}
            \State $E_{r_i,r_j}, \tilde{E}_{r_i}, \tilde{E}_{r_j},\gamma_{r_i,r_j}=$ None, None, None, None.
        \EndFor
        \For{$(i,j,r_i,r_j) \in \mathcal{B}$} \Comment{$\mathcal{O}(|\mathcal{B}|) \leq \mathcal{O}(n)$}
            \If{$\gamma_{r_i,r_j}$ is None }
                \State Calculate $\gamma_{r_i,r_j} $ according to Eq. 6
                \State $\gamma_{r_i,r_j} \gets \gamma_{r_j,r_i}$
            \EndIf
            \State Calculate $w_{r_i,r_j}$ according to Eq. 6
            \State $s \gets \gamma_{r_i,r_j}w_{r_i,r_j}$  \Comment{$\mathcal{O}(k_d \log{n})$}
            \State $E_{r_i,r_j}+=s$
            \State $E_{r_j,r_i} \gets E_{r_i,r_j}$
            \State $\tilde{E}_{r_i} \gets E_{r_i,r_j}$, if $E_{r_i,r_j}>\tilde{E}_{r_i}$
            \State $\tilde{E}_{r_j} \gets E_{r_i,r_j}$, if $E_{r_i,r_j}>\tilde{E}_{r_j}$
        \EndFor
        
        \For{$E_{i,j} \in E$} \Comment{$\mathcal{O}(|E|) \leq \mathcal{O}(|\mathcal{B}|)$}
            \If{$\frac{E_{i,j}}{\tilde{E}_i}<\alpha$ or $\frac{E_{i,j}}{\tilde{E}_j}<\alpha$}
                \State $E_{i,j} \gets 0$ 
            \EndIf
        \EndFor
    \end{algorithmic}
\end{algorithm}

For Algorithm 2, its time complexity is $\mathcal{O}(k_d |\mathcal{B}|\log{n})$. For each boundary point, the maximum number of corresponding boundary pair is $k_n$. Thus $ |E| \leq |\mathcal{B}|\leq k_n N_{boundary}\leq k_n n$, where $N_{boundary}$ is the number of boundary points.  Finally, the time complexity of Algorithm 2 is $\mathcal{O}(n\log{n})$.

\section*{A.2 Hyper-parameters used for experiments}

\begin{table*}[h]
\centering
\caption{Final hyper-parameters used for experiments on toy datasets}
\resizebox{\textwidth}{!}{
\begin{tabular}{llllll}
\toprule
name of parameters                          & Circles      & Moons         & Impossible  & S-set           & Smile \\
\toprule
GDT: $k_d$,$k_s$,$\alpha$,$\epsilon$              & 20,20,0.4,0  & 30,20,0.3,0   & 30,10,0.2,0 & 15,15,0.2,0     & 15,15,0.2,0 \\
HDBSCAN: min\_cluster\_size,min\_samples    & 10,10        & 2,11         & 11,5        & 20,10           & 20,10\\
Mean-Shift: quantile, n\_samples            & 0.2,500      & 0.1,500       & 0.2,500     & 0.1,500         & 0.3,500 \\
OPTICS: min\_samples, min\_cluster\_size    & 2,20         & 2,30          & 5,400       & 10,300          & 3,40  \\
Spectral Clustering: n\_clusters, affinity  & 2,"rbf"      & 2,"rbf"       & 6,"rbf"     & 15,"rbf"        & 4,"rbf" \\
k-means: n\_clusters                        & 2            & 2             & 6           & 15              & 4     \\
\toprule
\end{tabular}
}
\end{table*}

\begin{table*}[h]
\centering
\caption{Final hyper-parameters used for experiments on real-world datasets}
\resizebox{\textwidth}{!}{
\begin{tabular}{llllll}
\toprule
name of parameters                          & Iris          & Wine          & Cancer         & Glass             & Hepatitis \\
\toprule
GDT: $k_d$,$k_s$,$\alpha$,$\epsilon$              & 10,7,0.4,0  & 20,10,0.3,0   & 100,80,1,0           & 10,15,1,0.002     & 20,13,0,0 \\
HDBSCAN: min\_cluster\_size,min\_samples    & 30,20         & 20,2          & 10,10                 & 15,5              & 4,2\\
Mean-Shift: quantile,n\_samples             & 0.1,300       & 0.1,300       & 0.4,300               & 0.2,300           & 0.1,300\\
OPTICS: min\_samples,min\_cluster\_size     & 3,3           & 2,12          & 2,10                  & 3,8               & 2,2  \\
Spectral Clustering: n\_clusters,affinity   & 3,"rbf"       & 3,"rbf"       & 2,"rbf"               & 6,"rbf"           & 2,"rbf" \\
k-means: n\_clusters                        & 3             & 3             & 2                     & 6                 & 2     \\
\toprule
\end{tabular}
}
\end{table*}
\newpage
\section*{A.3 Analysis of hyper-parameters}
There are four parameters in DGSFC: $k_d$, $k_s$, $\alpha$ and $\epsilon$.

\paragraph{Selecting $k_d$.} $k_d$ is the neighborhood system's k in LKDE. A larger $k_d$ makes the density function smoother. In experiments, we usually choose $k_d$ from 10 to 30 in low-dimension case. In high-dimensional space, $k_d$ shall be larger.

\begin{figure}[H]
	\centering
		\subfigure[$k_d=5$]{ 
			\includegraphics[width=0.15\linewidth]{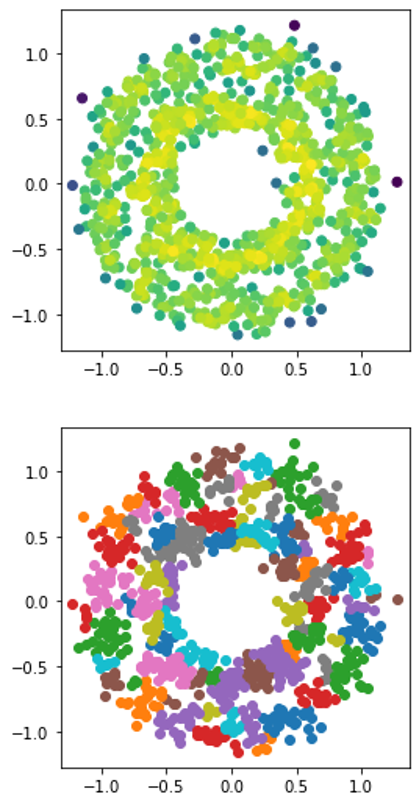}}
		\subfigure[$k_d=10$]{ 
			\includegraphics[width=0.15\linewidth]{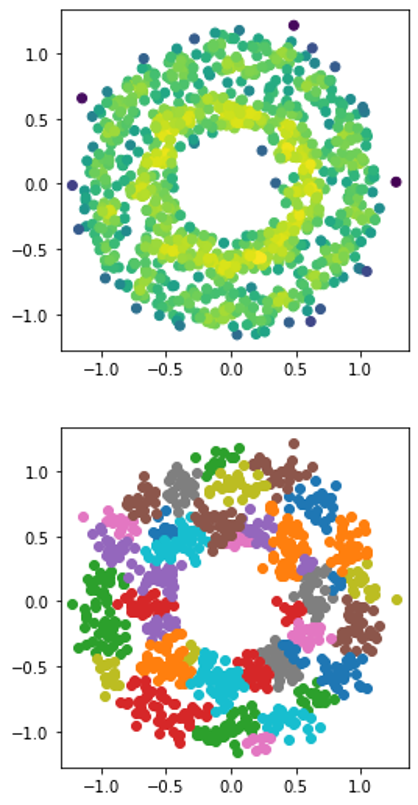}}
		\subfigure[$k_d=20$]{ 
			\includegraphics[width=0.15\linewidth]{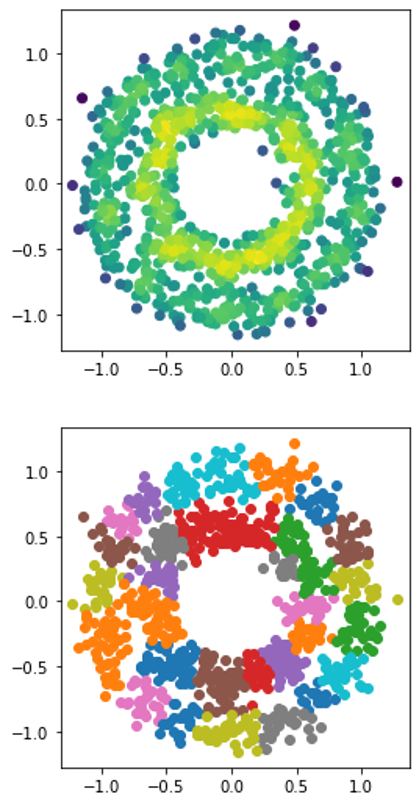}}
		\caption{ A larger $k_d$ leads to the smoother density function, and the small number of local clusters.}
		\label{figapp:change k_d}
\end{figure}

\paragraph{Selecting $k_s$.} $k_s$ is the neighborhood system's k for estimating the gradient direction for each point in density growing process. A smaller $k_s$ makes the local clusters more diverse. $k_s$ is usually smaller than $k_d$, we choose $k_d$ from 5 to 30 in low-dimensional case.

\begin{figure}[H]
	\centering
		\subfigure[$k_d=15$]{ 
			\includegraphics[width=0.15\linewidth]{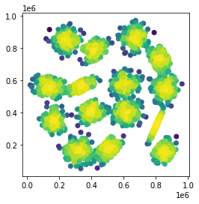}}
		\subfigure[$k_s=5$]{ 
			\includegraphics[width=0.15\linewidth]{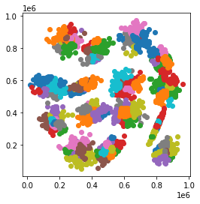}}
		\subfigure[$k_s=10$]{ 
			\includegraphics[width=0.15\linewidth]{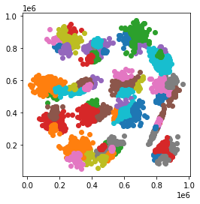}}
		\subfigure[$k_s=15$]{ 
			\includegraphics[width=0.15\linewidth]{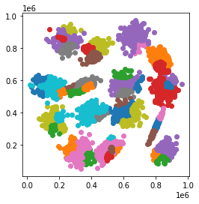}}
		\caption{ A smaller $k_s$ makes the local clusters more diverse. }
		\label{figapp:change k_s}
\end{figure}

\paragraph{Selecting $\alpha$.} $\alpha \in [0,1]$, affecting the threshold of preserving edges and equaling the $\lambda$ of the paper. A larger $\alpha$ results in a variety of final clusters. If there is only one label of the data points, the edges should not be pruned, and the established topo-graph will be a connected graph, forcing all the local clusters sharing one label. If the prior knowledge shows that there are lots of clusters, $\alpha$ shall be larger, and vise versa.

\begin{figure}[H]
	\centering
		\subfigure[local cluster]{ 
			\includegraphics[width=0.15\linewidth]{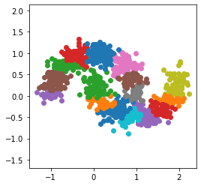}}
		\subfigure[$\alpha=0$]{ 
			\includegraphics[width=0.15\linewidth]{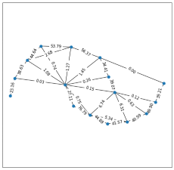}}
		\subfigure[$\alpha=0.1$]{ 
			\includegraphics[width=0.15\linewidth]{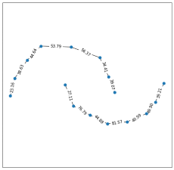}}
		\subfigure[$\alpha=0.2$]{ 
			\includegraphics[width=0.15\linewidth]{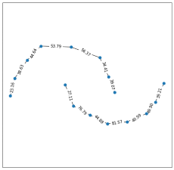}}
		\subfigure[$\alpha=0.5$]{ 
			\includegraphics[width=0.15\linewidth]{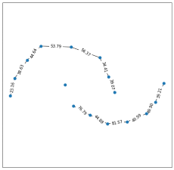}}
		\caption{ A larger $\alpha$ ensures that the final clusters are more diverse. }
		\label{figapp:change alpha}
\end{figure}

\paragraph{Selecting $\epsilon$.} $\epsilon \in [0,1]$, affecting the threshold for judging if a sample point should be taken as a noise point. A larger $ \epsilon $ helps to detect more noise points. Unless there are huge noise points hurting the results significantly, $ \epsilon $ is chosen as small as possible, because we tend to specify a potential cluster for each sample. The comparative advantage is that algorithms like HDBSCAN may drop too many points as 'noise', resulting in poor performance on cover rate and the  ability of discovering topological structure of the dataset.

\begin{figure}[H]
	\centering
		\subfigure[local cluster]{ 
			\includegraphics[width=0.2\linewidth]{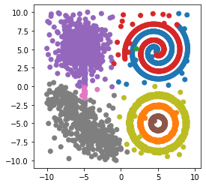}}
		\subfigure[$\alpha=0$]{ 
			\includegraphics[width=0.2\linewidth]{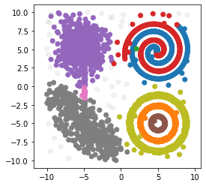}}
		\subfigure[$\alpha=0.1$]{ 
			\includegraphics[width=0.2\linewidth]{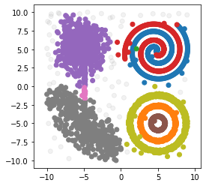}}
		\subfigure[$\alpha=0.2$]{ 
			\includegraphics[width=0.2\linewidth]{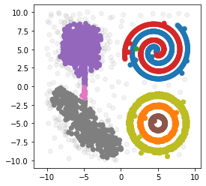}}
		\caption{ A larger $ \epsilon $ helps to detect more noise points. }
		\label{figapp:change epsilon}
\end{figure}
\newpage
\section*{A.4 Visualization of toy datasets}
\begin{figure*}[h]
	\centering
	    \includegraphics[width=5.5in]{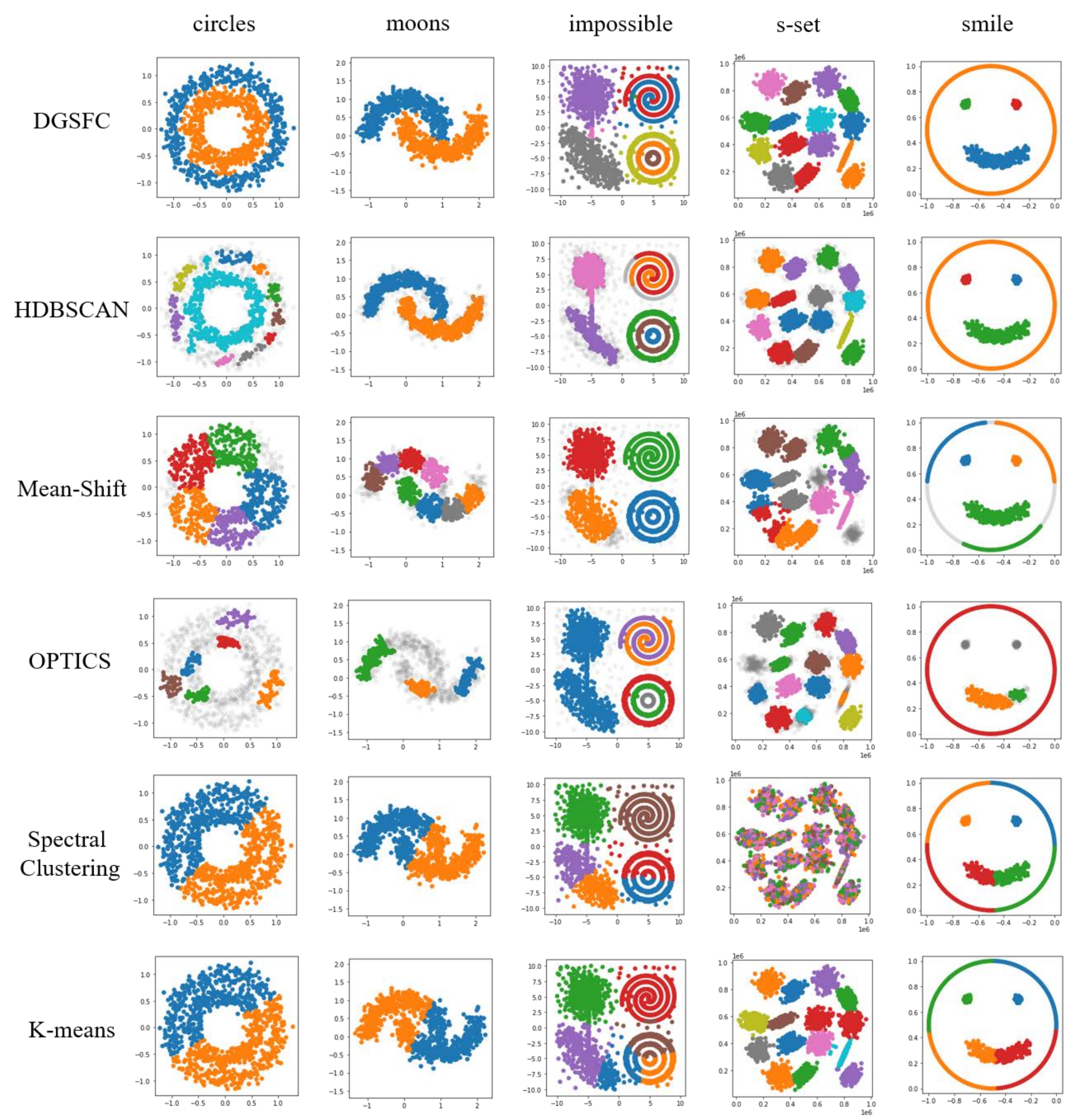}
		\caption{ Visualization of results of different algorithms on toy datasets: rows and columns represent different algorithms and datasets respectively. }
		\label{figapp:boundary points}
\end{figure*}
\newpage
\section*{A.5 Visualization of segmentation}
\begin{table*}[h]
\centering
\caption{Final hyper-parameters used for experiments of segmentation task}
\resizebox{\textwidth}{!}{
\begin{tabular}{lllllllll}
\toprule
name of parameters                          & Red House          & Westlake Night          & Westlake Daytime         & Westlake Twilight             & Westlake Pavilion          & Westlake Temple\\
\toprule
GDT: $k_d$,$k_s$,$\alpha$,$\epsilon$        & 30,20,0.05,0.0001  & 40,10,0.15,0          & 40,15,0.1,0     & 40,8,0.05,0     &40,20,0.03,0       & 40,20,0.08,0\\
HDBSCAN: min\_cluster\_size,min\_samples    & 16,20         & 30,10          & 10,10                 & 30,10              & 30,5          & 30,2\\
\toprule
\end{tabular}
}
\end{table*}

\begin{figure*}[h]
	\centering
	    \includegraphics[width=6 in]{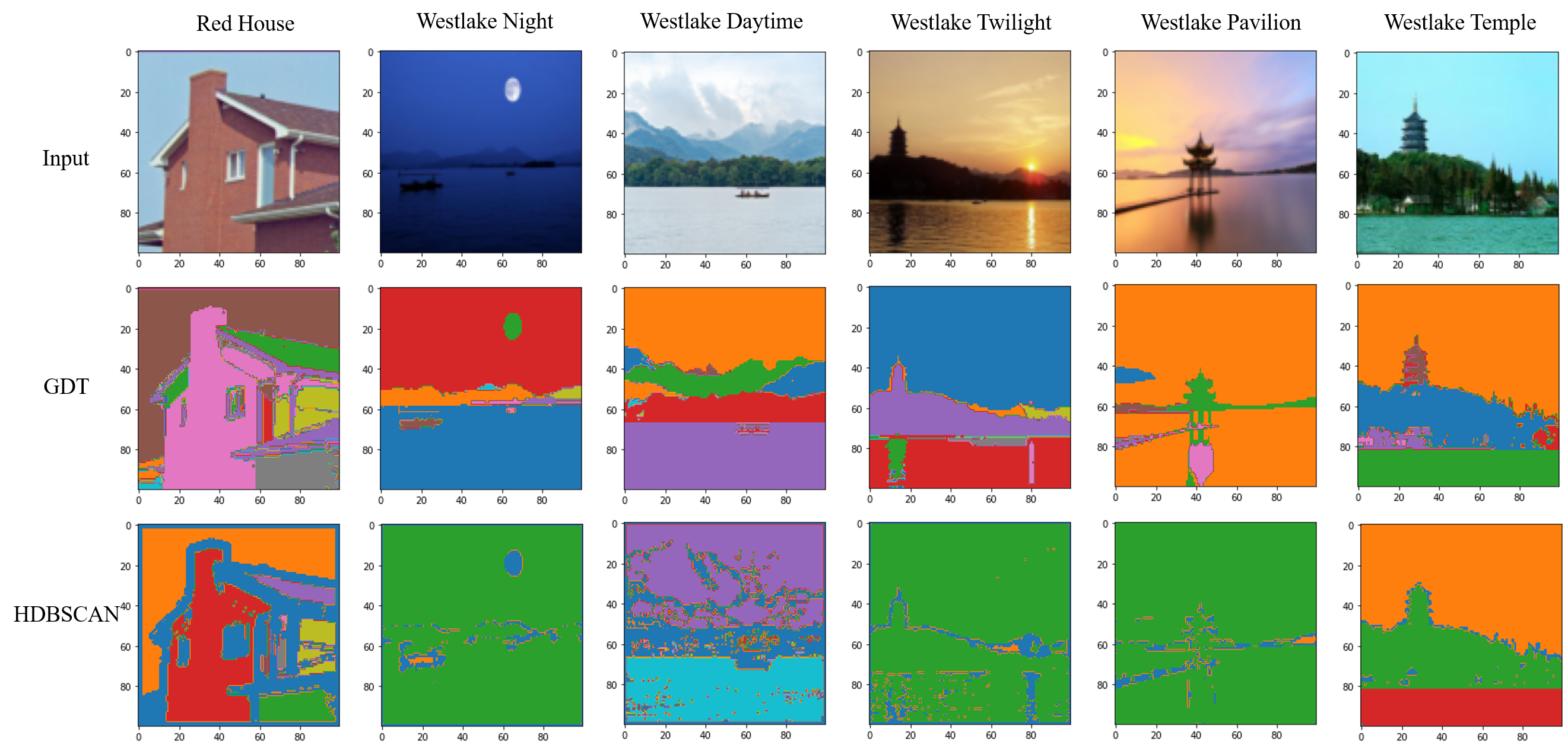}
		\caption{ Visualization of results of different algorithms on image segmentation tasks: rows and columns represent different algorithms and datasets respectively. }
		\label{figapp:segmentation}
\end{figure*}

\section*{A.6 Computing Infrastructure}

\begin{table*}[h]
\centering
\caption{The computing infrastructure}
\resizebox{\textwidth}{!}{
\begin{tabular}{lllll}
\toprule
                           CPU          & amount of memory          & operating system    & version of python      & version of libraries\\
\toprule
       64  AMD Ryzen Threadripper 3970X 32-Core Processor  & 251.84GB          & Ubuntu 18.04.3 LTS   & 3.6.10   & seeing the README.md of the code\\
\toprule
\end{tabular}
}
\end{table*}

\end{document}